%% file: main.tex
\documentclass[10pt,twocolumn,letterpaper]{article}

\usepackage{iccv}              

\input{preamble}

\definecolor{iccvblue}{rgb}{0.21,0.49,0.74}
\usepackage[pagebackref,breaklinks,colorlinks,allcolors=iccvblue]{hyperref}

\def\paperID{5375} 
\def\confName{ICCV}
\def\confYear{2025}

\title{HiNeuS: High-fidelity Neural Surface \\ Mitigating Low-texture and Reflective Ambiguity}

\author{Yida Wang \and Xueyang Zhang \and Kun Zhan \\ \\ Li Auto Inc.\\ \and Peng Jia \and Xianpeng Lang 
}

\begin{document}

\makeatletter
\g@addto@macro\@maketitle{
\begin{figure}[H]
    \setlength{\linewidth}{\textwidth}
    \setlength{\hsize}{\textwidth}
    \centering
    \includegraphics[width=\linewidth]
{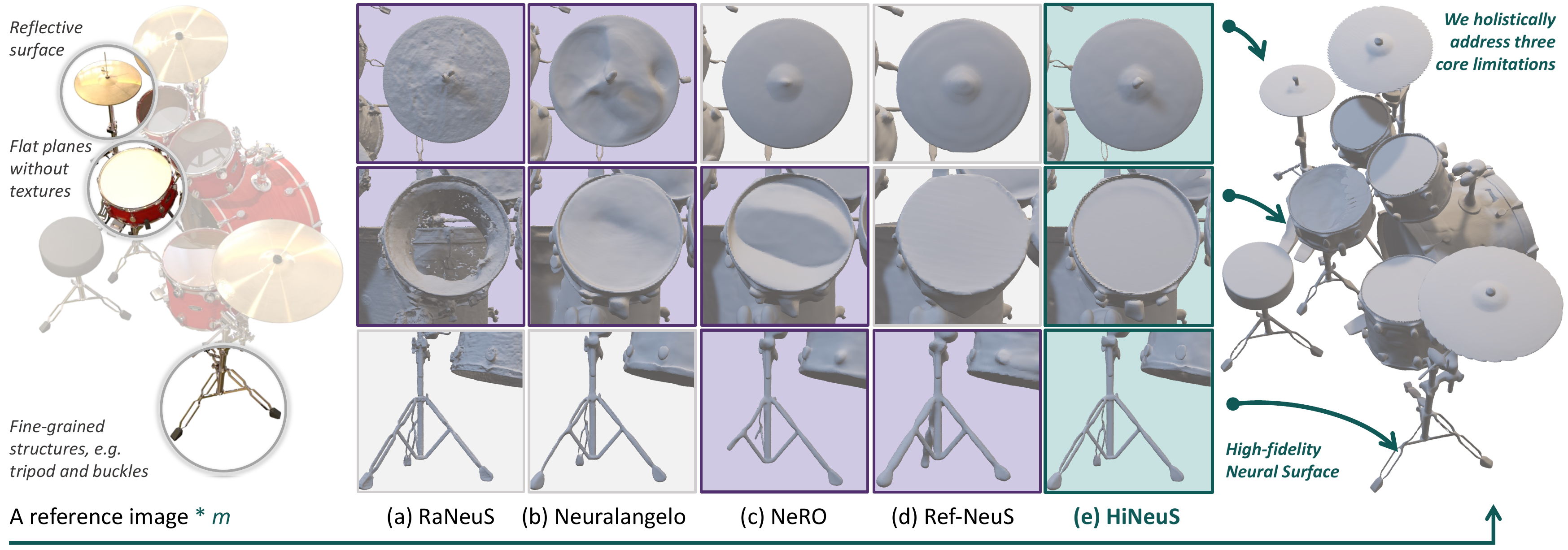}
\caption{Structural comparison among approaches targeting learning neural surfaces towards a set of drums. Our proposed method delivers the best performance on reconstructing 1) reflective structures, 2) low-textured surfaces, and 3) fine-grained geometries.}
\label{fig:teaser}
\end{figure}
}

\maketitle

\input{sec/0_abstract}    
\input{sec/1_intro}
\input{sec/2_method}
\input{sec/3_exp}
{
    \small
    \bibliographystyle{ieeenat_fullname}
    \bibliography{main}
}
\input{sec/X_suppl}

\end{document}

%% file: preamble.tex
\usepackage{graphicx}
\usepackage{booktabs}
\usepackage{graphicx}%
\usepackage{multirow}%
\usepackage{amsmath,amssymb,amsfonts}%
\usepackage{amsthm}%
\usepackage{mathrsfs}%
\usepackage[title]{appendix}%
\usepackage{xcolor}%
\usepackage{textcomp}%
\usepackage{manyfoot}%
\usepackage{booktabs}%
\usepackage{algorithm}%
\usepackage{algorithmicx}%
\usepackage{algpseudocode}%
\usepackage{listings}%
\usepackage{physics}
\usepackage{tikz} 
\usepackage{bm}

\definecolor{liautogreen}{RGB}{0, 85, 80}

\newcommand{\figref}[1]{Fig.~\ref{#1}}
\newcommand{\secref}[1]{Sec.~\ref{#1}}
\newcommand{\tabref}[1]{Table~\ref{#1}}


%% file: sec/0_abstract.tex
\begin{abstract}
Neural surface reconstruction faces persistent challenges in reconciling geometric fidelity with photometric consistency under complex scene conditions. We present HiNeuS, a unified framework that holistically addresses three core limitations in existing approaches: multi-view radiance inconsistency, missing keypoints in textureless regions, and structural degradation from over-enforced Eikonal constraints during joint optimization. To resolve these issues through a unified pipeline, we introduce: 1) Differential visibility verification through SDF-guided ray tracing, resolving reflection ambiguities via continuous occlusion modeling; 2) Planar-conformal regularization via ray-aligned geometry patches that enforce local surface coherence while preserving sharp edges through adaptive appearance weighting; and 3) Physically-grounded Eikonal relaxation that dynamically modulates geometric constraints based on local radiance gradients, enabling detail preservation without sacrificing global regularity. Unlike prior methods that handle these aspects through sequential optimizations or isolated modules, our approach achieves cohesive integration where appearance-geometry constraints evolve synergistically throughout training. Comprehensive evaluations across synthetic and real-world datasets demonstrate state-of-the-art performance, including a 21.4\% reduction in Chamfer distance over reflection-aware baselines and 2.32 dB PSNR improvement against neural rendering counterparts. Qualitative analyses reveal superior capability in recovering specular instruments, urban layouts with centimeter-scale infrastructure, and low-textured surfaces without local patch collapse. The method's generalizability is further validated through successful application to inverse rendering tasks, including material decomposition and view-consistent relighting. \textcolor{liautogreen}{\textbf{Codes will be publicly available.}}

\end{abstract}

%% file: sec/1_intro.tex
\section{Introduction}
\label{sec:intro}

Neural surface reconstruction has emerged as a pivotal technique in 3D computer vision, enabling high-fidelity geometry recovery through multi-view supervision~\cite{mildenhall2021nerf, yariv2021volume}. While neural rendering approaches like NeRF~\cite{mildenhall2021nerf} achieve photorealistic view synthesis, their implicit density fields often fail to recover precise surfaces~\cite{wang2021neus}. Recent advances in signed distance function (SDF)-based methods~\cite{yariv2021volume,wang2021neus} bridge this gap by integrating geometric priors with volume rendering. However, as illustrated in \figref{fig:teaser}, reconstructing surfaces with reflective materials, textureless regions, and fine details remains challenging due to inherent multi-view ambiguities and optimization conflicts.

Existing methods often address these challenges in isolation. For instance, reflection-aware techniques like Ref-NeuS~\cite{ge2023ref} handle specular surfaces but introduce noise in low-textured areas (\secref{ssec:self_reflection}). Surface regularization approaches~\cite{wang2022hf} improve planar regions but oversmooth geometric details. Meanwhile, high-fidelity methods like Neuralangelo~\cite{li2023neuralangelo} recover intricate structures but struggle with view-dependent effects. This fragmented progress leaves three fundamental challenges unresolved:

\begin{itemize}
    \item \textbf{Multi-view inconsistency}: Strong reflections and indirect illumination violate photometric consistency assumptions, causing geometric artifacts (Fig.~\ref{fig:toaster}).
    \item \textbf{Low-textured surfaces}: Sparse visual cues lead to over-regularization, eroding valid structures (\secref{ssec:local_geom}). 
    \item \textbf{Detail-geometry conflict}: Traditional Eikonal constraints~\cite{gropp2020implicit} prioritize smoothness over high-frequency details, as shown in \figref{fig:eikonal_relax}.
\end{itemize}

We present a unified framework that simultaneously addresses these challenges through three key innovations:

\begin{enumerate}
    \item \textbf{SDF-guided multi-view consistency}: Leveraging continuous SDF evaluation (\secref{ssec:mv_consistency}), we compute visibility factors $\mathbb{V}_j$ (Eq.~\ref{equ:sdf_visibility}) to resolve reflection ambiguities without mesh discretization artifacts.
    \item \textbf{Local geometry-constrained regularization}: Our ray-aligned planar constraints (\secref{ssec:local_geom}) adaptively regularize textureless regions while preserving edges through feature-aware weighting $\lambda_\text{pla}^k$.
    \item \textbf{Rendering-prioritized Eikonal relaxation}: An adaptive weighting scheme $\omega(\bm{x})$ (Eq.~\ref{equ:adaptive_weight}) dynamically balances geometric fidelity and rendering accuracy, enabling detail preservation in high-error regions.
\end{enumerate}

Our experiments across synthetic and real-world datasets (\secref{sec:ablation}) demonstrate state-of-the-art performance, achieving a 21.4\% improvement in Chamfer distance over Ref-NeuS~\cite{ge2023ref} on reflective surfaces (Table~\ref{tab:glossy_nero}) and 35.00 dB PSNR on NeRF-Synthetic (Table~\ref{tab:psnr_synthetic}). The ablation studies confirm that our components synergistically address the three challenges without performance trade-offs.

\section{Related Works}
\label{sec:related}

\subsection{Neural Surface Reconstruction}
The evolution of neural surface reconstruction builds upon two foundational paradigms: multi-view stereo (MVS) geometry recovery~\cite{schoenberger2016mvs} and neural radiance field rendering~\cite{mildenhall2021nerf}. While traditional MVS pipelines like COLMAP~\cite{schoenberger2016sfm} establish photogrammetric baselines, neural rendering approaches achieve unprecedented view synthesis quality through continuous volumetric representations~\cite{barron2021mip,zhang2020nerf++}. Subsequent works address computational bottlenecks via hash encoding strategies~\cite{muller2022instant,wang2022neus2}, enabling real-time performance but sacrificing geometric precision. Our method bridges this gap by maintaining rendering-quality supervision while enforcing physically-grounded surface constraints.

\subsection{Implicit Surface Representations}
Modern neural reconstruction systems predominantly employ signed distance fields (SDF) due to their mathematical surface definition $\mathcal{M} = \{\bm{x} \in \mathbb{R}^3 | f(\bm{x}) = 0\}$. Seminal works like VolSDF~\cite{yariv2021volume} and NeuS~\cite{wang2021neus} establish differentiable SDF-to-density mappings using Laplace and logistic distributions respectively. While these enable watertight surface extraction through volume rendering supervision, they suffer from three key limitations our method addresses: 1) Reflection-induced multi-view inconsistencies, 2) Over-smoothing in textureless regions, and 3) Detail erosion from uniform Eikonal constraints.

Recent advances tackle specific aspects of these challenges. Geo-NeuS~\cite{fu2022geo} incorporates MVS depth priors but struggles with specular surfaces. HF-NeuS~\cite{wang2022hf} employs hierarchical feature grids yet produces artifacts in low-texture areas. NeuralUDF~\cite{long2022neuraludf} extends to arbitrary topologies via unsigned distance fields (UDF) but requires dense view sampling. Our unified framework transcends these limitations through three innovations: SDF-grounded visibility verification, adaptive local regularization, and rendering-conditioned geometric constraints.

\subsection{Reflection-Aware Reconstruction}
Handling specular surfaces remains challenging due to view-dependent radiance violating multi-view consistency. Neural approaches like Ref-NeuS~\cite{ge2023ref} model reflections through parametric BRDFs but fail on complex real-world materials. Mesh-based methods~\cite{liu2023nero} suffer from tessellation artifacts in thin structures. Our SDF-guided visibility factor $\mathbb{V}$ (Section~\ref{ssec:mv_consistency}) overcomes these limitations through continuous occlusion reasoning without surface discretization. Unlike NeRO's learned reflection probabilities~\cite{liu2023nero}, our physics-inspired formulation (Eq.~\ref{eq:indirect_reflect}) maintains differentiability while preserving thin structures.

\subsection{Geometry-Appearance Co-Regularization}
Balancing surface smoothness with detail preservation constitutes a fundamental challenge. Traditional approaches apply uniform Eikonal constraints~\cite{gropp2020implicit} or TV regularization~\cite{izadi2011kinectfusion}, often eroding fine structures. NeuralWarp~\cite{darmon2022improving} enforces photometric consistency but struggles with textureless regions. Our method introduces two key advances: 1) Local planar constraints $\mathbb{P}$ (Section~\ref{ssec:local_geom}) that adapt to radiance variations, and 2) Rendering-prioritized Eikonal relaxation (Eq.~\ref{equ:relaxed_eikonal}) that dynamically weights geometric constraints based on reconstruction error. This dual mechanism preserves details while preventing over-regularization, outperforming both SparseNeuS~\cite{long2022sparseneus} and Instant-NSR~\cite{zhao2022human} in complex scenes (Table~\ref{tab:psnr_synthetic}).

%% file: sec/2_method.tex
\section{Methodology}
\label{sec:method}

Our method bridges neural rendering with geometric reconstruction through three key innovations: 1) SDF-guided multi-view consistency verification, 2) Local geometry-constrained textureless surface regularization, and 3) Rendering-prioritized Eikonal relaxation. \figref{fig:teaser} overviews our framework.

\subsection{Neural Surface-Supervised Rendering}
\label{ssec:neural_rendering}
Building on volume rendering fundamentals~\cite{mildenhall2021nerf}, we render color $\bm{C}(\bm{r})$ for ray $\bm{r}$ with origin $\bm{o}$ and direction $\bm{v}$ as
\begin{equation}
    \bm{C}(\bm{r}) = \int_{t_n}^{t_f} T(t)\sigma(\bm{r}(t))\bm{c}(\bm{r}(t),\bm{v})\,dt ~,
    \label{equ:radiance}
\end{equation}
where transparency $T(t) = \exp\left(-\int_{t_n}^t \sigma(\bm{r}(s))\,ds\right)$. Unlike NeRF's density field, we derive $\sigma$ from the SDF $f$ via
\begin{equation}
    \sigma(t) = \alpha\Psi_s(-f(\bm{r}(t))) ~,
    \label{equ:density}
\end{equation}
where $\Psi_s$ is the Laplace cumulative distribution function (CDF)~\cite{yariv2021volume} with learnable scale $s$, and $\alpha$ controls density decay. This SDF-to-density mapping enables surface extraction via $\mathcal{M} = \{\bm{x} \in \mathbb{R}^3 \mid f(\bm{x}) = 0\}$.

\subsection{SDF-Guided Multi-View Consistency}
\label{ssec:mv_consistency}

\begin{figure}[t]
    \centering
    \includegraphics[width=0.99\linewidth]{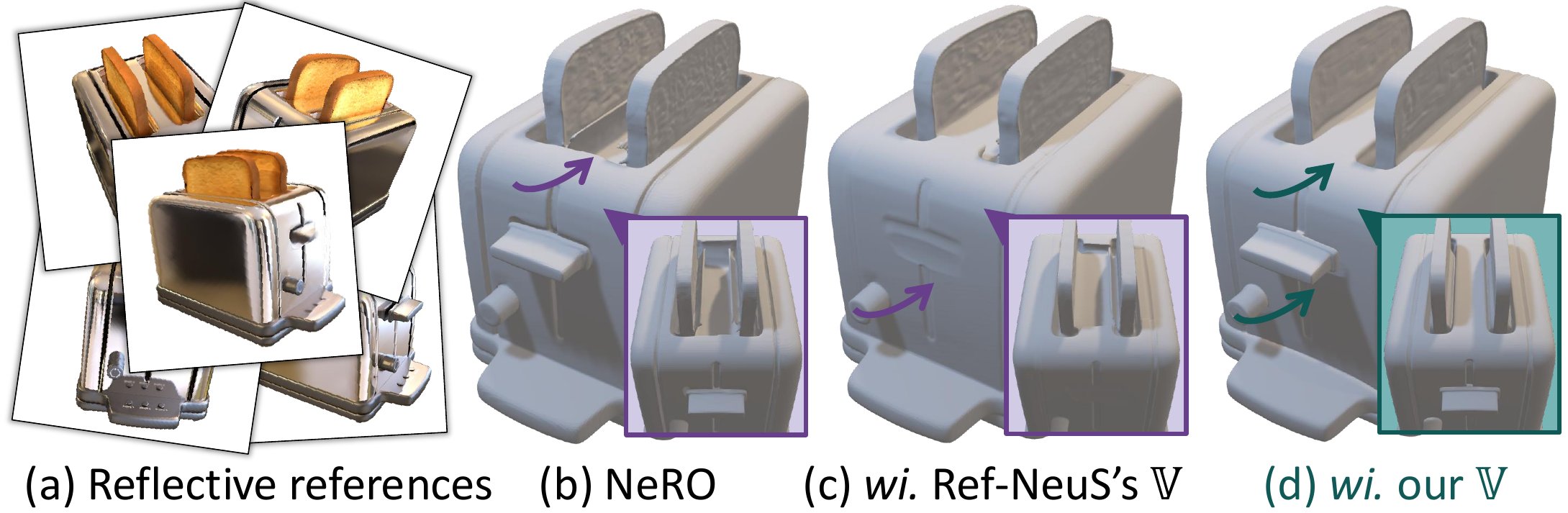}
    \caption{Occlusion-aware reflection handling. Our SDF-based visibility verification (d) avoids artifacts in neural (b) and mesh-based (c) approaches through continuous surface evaluation.}
    \label{fig:toaster}
\end{figure}

Given the optimization loss $\mathcal{L}_\text{rgb} = \frac{1}{m}\sum_{i=1}^m \|\bm{\hat{C}}_i - \bm{C}_i\|_2$, the supervision $\bm{\hat{C}}_i$ might be with ambiguity due to strong reflection observed from ray $\bm{r}_i$. To mitigate such ambiguity, we first introduce an ambiguity factor $\lambda_\text{ambiguity}(\bm{r})$ quantifies multi-view consistency for ray $\bm{r}$ using geometric and photometric constraints
\begin{align}
    \lambda_\text{ambiguity}(\bm{r}) &= \frac{1}{|\mathcal{V}_v|}\sum_{j\in\mathcal{V}_v} \mathbb{V}_j D_M(\bm{C}_i, \bm{C}_j) \nonumber \\
    D_M(\bm{C}_i, \bm{C}_j) &= \sqrt{(\bm{C}_i - \bm{C}_j)^\top\bm{\Sigma}^{-1}(\bm{C}_i - \bm{C}_j)} ~,
    \label{equ:ambiguity_factor}
\end{align}
where $\mathcal{V}_v$ denotes views with $\mathbb{V}_j > 0.9$ (visible views), $\bm{\Sigma}$ is the empirical covariance matrix computed across training images, and $D_M$ is the Mahalanobis distance in RGB space. The visibility factor $\mathbb{V}_j$ from view $j$ given surface point $\bm{x}_0^i$ in view $i$ is computed through continuous SDF evaluation:
\begin{align}
    \mathbb{V}_j &= \prod_{k=1}^K \sigma\left(\beta f(\bm{x}_k^{(j)})\right) ~,
    \label{equ:sdf_visibility}
\end{align}
where $\sigma(z) = (1 + e^{-z})^{-1}$ is the sigmoid function with sharpness $\beta > 0$, and $\bm{x}_k^{(j)}$ are samples along the ray from $\bm{o}_j$ to $\bm{x}_0^i$:
\begin{align}
    \bm{r}_j(t) &= \bm{o}_j + t\frac{\bm{x}_0^i - \bm{o}_j}{\|\bm{x}_0^i - \bm{o}_j\|}, \quad t \in [0, \|\bm{x}_0^i - \bm{o}_j\|] ~.
\end{align}

\paragraph{Physical interpretation.} $\sigma(\beta f(\bm{x})) \approx 1$ when $f(\bm{x}) > 0$ (free space), and $\sigma(\beta f(\bm{x})) \approx 0$ when $f(\bm{x}) < 0$ (occupied space) in \eqref{equ:sdf_visibility}. The product over ray samples implies visibility $\mathbb{V}_j \approx 1$ if the entire ray remains in free space. This formulation provides three key advantages: (i) No tessellation artifacts, (ii) Occlusion detection behind thin structures, and (iii) Photometric constraints only for visible points.

\begin{algorithm}[t]
\caption{SDF Visibility Verification}
\begin{algorithmic}[1]
\Require Camera centers $\{\bm{o}_j\}_{j=1}^N$, surface point $\bm{x}_0^i$
\Ensure Visibility factors $\{\mathbb{V}_j\}_{j=1}^N$
\For{each view $j \in \{1,...,N\}$}
    \State $\bm{v}_j \leftarrow (\bm{x}_0^i - \bm{o}_j)/\|\bm{x}_0^i - \bm{o}_j\|$
    \State Sample $\{t_k\} \sim \text{Uniform}(0, \|\bm{x}_0^i - \bm{o}_j\|)$
    \State $\mathbb{V}_j \leftarrow 1.0$
    \For{each sample $t_k$}
        \State $\bm{x}_k \leftarrow \bm{o}_j + t_k\bm{v}_j$
        \State $\mathbb{V}_j \leftarrow \mathbb{V}_j \cdot \sigma(\beta f(\bm{x}_k))$
    \EndFor
\EndFor
\end{algorithmic}
\end{algorithm}
The ambiguity factor is integrated into training through loss weighting
\begin{align}
    \mathcal{L}_\text{rgb} = \frac{1}{m}\sum_{i=1}^m \frac{\|\bm{\hat{C}}_i - \bm{C}_i\|_2}{1 + \lambda_\text{ambiguity}(\bm{r}_i)} ~.
    \label{equ:loss_without_ambiguity}
\end{align}

\subsection{Self-Reflection Compensation} 
\label{ssec:self_reflection}

\begin{figure}[t]
    \centering
    \includegraphics[width=0.99\linewidth]{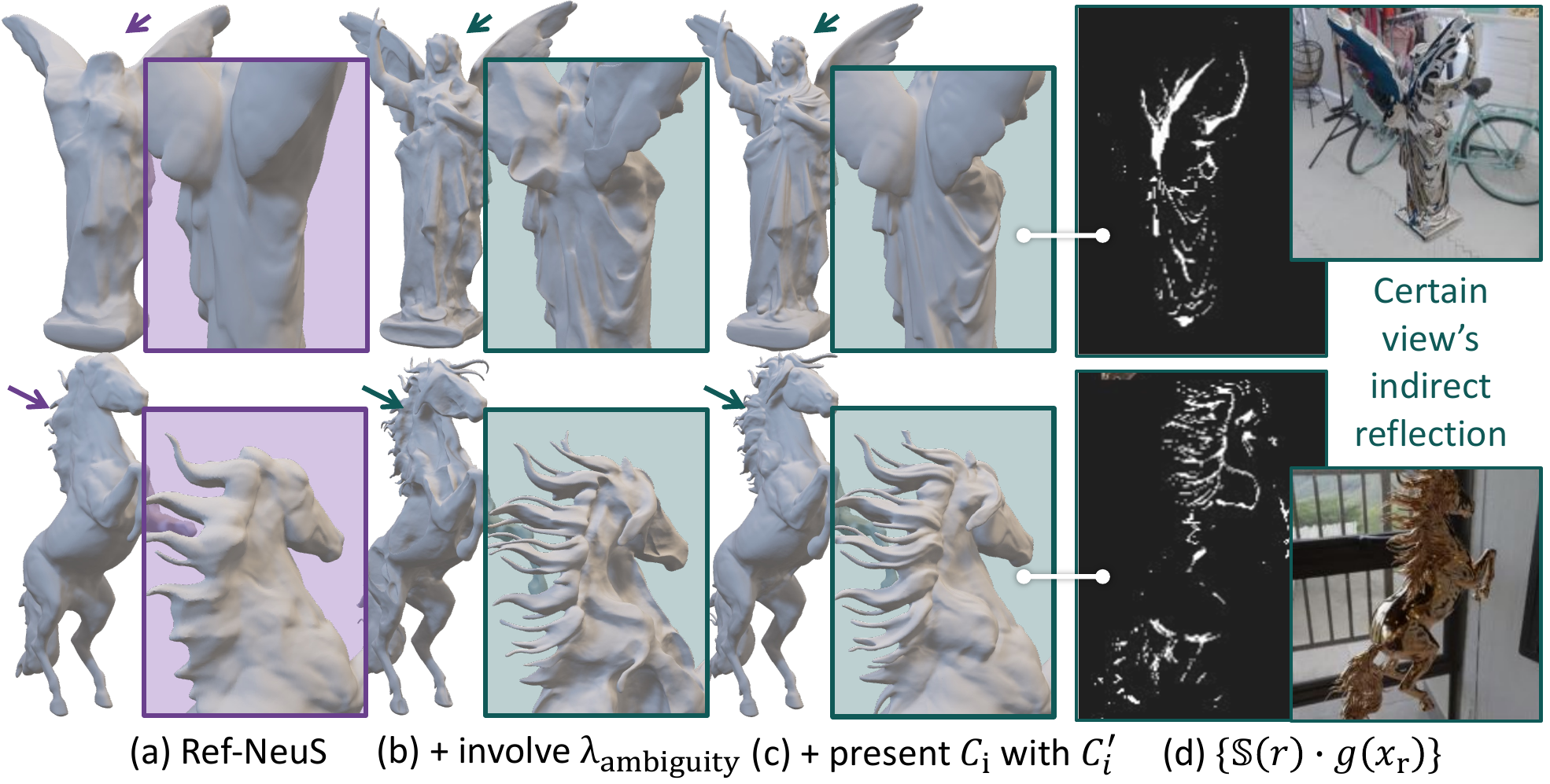}
    \caption{Self-reflection handling comparison: (a) RefNeuS~\cite{ge2023ref} - trained with open-sourced codes, (b) Trained by \eqref{equ:loss_without_ambiguity} with ambiguity factor, (c) Trained with \eqref{eq:reflection_loss}, by reformating $\bm{C}$ as $\bm{C}'$, an exemplar indirect compensation $\mathbb{S}(\bm{r})g(\bm{x}_r)$ is shown on the right in (c).}
    \label{fig:self_reflect}
\end{figure}

For rays exhibiting indirect reflections, we model the radiance as $\bm{C}'(\bm{r}) = (1 - \mathbb{S}(\bm{r}))\bm{C}(\bm{r}) + \mathbb{S}(\bm{r})g(\bm{x}_r)$,
where $\mathbb{S}(\bm{r}) \in [0,1]$ is the reflection probability, and $\bm{x}_r$ is the secondary surface intersection. The probability is computed via
\begin{equation}
    \mathbb{S}(\bm{r}) = \max_{t \in [0,t_{\max}]} \Psi_{\gamma}(-f(\bm{r}_\text{refl}(t)))
    \label{eq:indirect_reflect}
\end{equation}
with Laplace CDF $\Psi_{\gamma}$ and the reflected ray
\begin{align}
    \bm{r}_\text{refl}(t) &= \bm{x}_0 + t\left(\bm{v} - 2(\bm{v} \cdot \bm{n})\bm{n}\right) \nonumber \\
    \bm{n} &= \nabla f(\bm{x}_0)/\|\nabla f(\bm{x}_0)\| ~,
\end{align}
where $\bm{x}_0$ is the primary surface point from ~\secref{ssec:neural_rendering}, and $t_{\max}$ is empirically set to be 0.1. The reflection MLP $g$ processes the secondary point $g(\bm{x}_r) = \text{MLP}(\bm{x}_r, \bm{n}_r, \bm{v}_r)$,
where $\bm{n}_r = \nabla f(\bm{x}_r)$ and $\bm{v}_r$ is the reflection direction.

\begin{algorithm}[t]
\caption{Indirect Reflection Probability}
\begin{algorithmic}[1]
\State Input: Primary ray $\bm{r}$ with surface hit $\bm{x}_0$, view dir $\bm{v}$
\State Compute reflection dir $\bm{v}_\text{refl} \gets \bm{v} - 2(\bm{v} \cdot \bm{n})\bm{n}$
\State March reflected ray $\bm{r}_\text{refl}(t) \gets \bm{x}_0 + t\bm{v}_\text{refl}$
\State Sample $\{t_k\} \sim \text{Uniform}(0, t_{\max})$
\State Compute $\mathbb{S}(\bm{r}) \gets \max_k \Psi_\gamma(-f(\bm{r}_\text{refl}(t_k)))$
\State Find $\bm{x}_r \gets \arg\min_k |f(\bm{r}_\text{refl}(t_k))|$
\State \Return $\mathbb{S}(\bm{r}), g(\bm{x}_r)$
\end{algorithmic}
\end{algorithm}

Compared to NeRO's learned reflections~\cite{liu2023nero}, our formulation physically constrains paths through the SDF $f$ while maintaining differentiability. This avoids discretization artifacts from mesh extraction (unlike NeRO's MLP-learned $\mathbb{S}$) while preserving thin structures.
Then the reflection-aware loss becomes
\begin{align}
    \mathcal{L}_\text{rgb} = \frac{1}{m}\sum_{i=1}^m \frac{\|\bm{\hat{C}}_i - \bm{C}'_i\|_2}{1 + \lambda_\text{ambiguity}(\bm{r}_i)} ~.
    \label{eq:reflection_loss}
\end{align}

\subsection{Local Geometry-Constrained Regularization}
\label{ssec:local_geom}

\begin{figure}[t]
    \centering
    \includegraphics[width=0.99\linewidth]{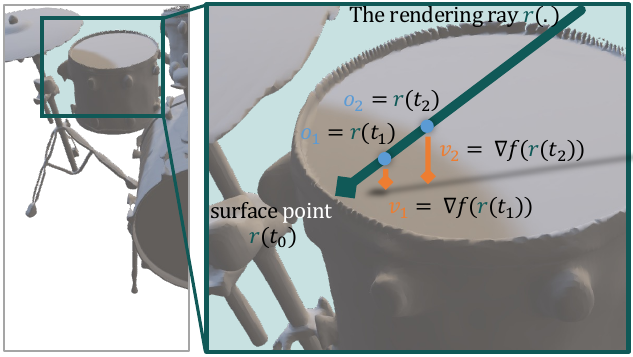}
    \caption{Local geometry constraints enforce planarity through ray-aligned neighborhood sampling in textureless regions.}
    \label{fig:smooth}
\end{figure}

For textureless regions, we enforce local smoothness through ray-constrained sampling. Given surface point $\bm{x}_0$ from Section~\ref{ssec:neural_rendering} on ray $\bm{r}(t) = \bm{o} + t\bm{v}$, sample $K$ neighboring points $
    \bm{x}_k = \bm{r}(t_0 + \Delta t_k), \quad \Delta t_k \sim \mathcal{U}(-\eta, 0)$,
where $t_0$ is the ray depth at $\bm{x}_0$, and $\eta$ controls the local sampling radius (Fig.~\ref{fig:smooth}). We enforce SDF linearity via:
\begin{equation}
    \mathcal{L}_\text{planar} = \frac{1}{K}\sum_{k=1}^K \lambda_\text{pla}^k\left|\frac{f(\bm{x}_k)}{\|\bm{x}_k - \bm{x}_0\|} - \bm{n}_0^\top\frac{\bm{x}_k - \bm{x}_0}{\|\bm{x}_k - \bm{x}_0\|}\right| ~,
\end{equation}
where $\bm{n}_0 = \nabla f(\bm{x}_0)/\|\nabla f(\bm{x}_0)\|$ is the unit normal. Let $\bm{c}_\text{feat}(\bm{x}_k)$ denote the radiance features observed from point $\bm{x}_k$ toward the zero-crossing surface location $\bm{x}_s$ where $f(\bm{x}_s)=0$, computed as
\begin{equation}
    \bm{c}_\text{feat}(\bm{x}_k) = \text{MLP}(\bm{x}_k, \bm{v}_k), \quad \bm{v}_k = \frac{\bm{x}_s - \bm{x}_k}{\|\bm{x}_s - \bm{x}_k\|}
    \label{eq:radiance_feat}
\end{equation}
The adaptive weighting uses a small constant $\epsilon = 10^{-3}$ for numerical stability $
    \lambda_\text{pla}^k = \frac{\epsilon}{\|\bm{c}_\text{feat}(\bm{x}_k) - \bm{c}_\text{feat}(\bm{x}_0)\|_2 + \epsilon}$.
This constraint preserves sharp edges while preventing artifacts in textureless regions.

\subsection{Rendering-Prioritized Eikonal Relaxation}
\label{ssec:eikonal_relax}

\begin{figure}[t]
    \centering
    \includegraphics[width=0.99\linewidth]{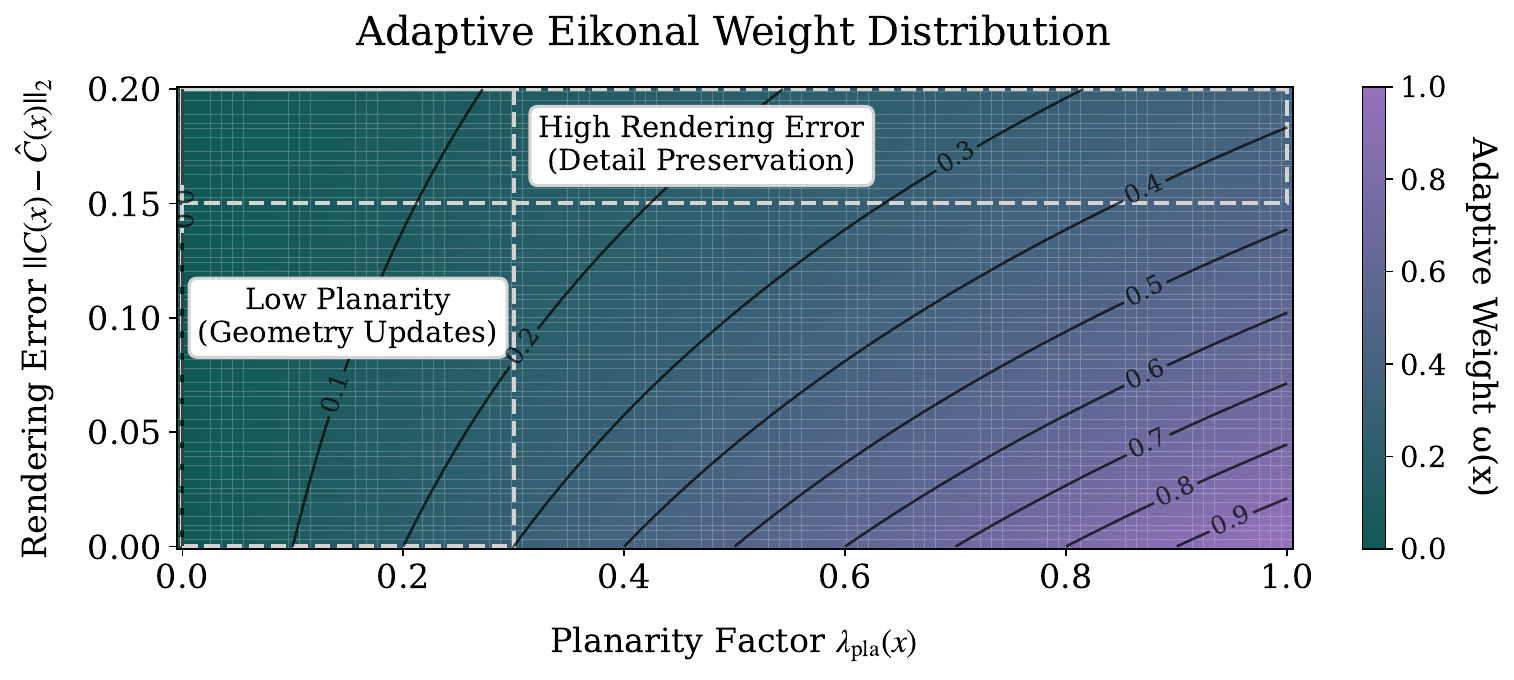}
    \caption{Adaptive weights $\omega(\bm{x})$ with rendering error (x-axis) and planarity factor $\lambda_\text{pla}$ (y-axis). Purple/green regions indicate strong/weak regularization respectively.}
    \label{fig:eikonal_relax}
\end{figure}

Our adaptive Eikonal regularization balances geometric fidelity and rendering accuracy through error-driven relaxation. Building on the neural rendering formulation from ~\eqref{equ:radiance} and density derivation in ~\eqref{equ:density}, we introduce bidirectional dependency between SDF values and rendering errors. The adaptive weight $\omega(\bm{x})$ relaxes Eikonal constraints
\begin{align}
    \omega(\bm{x}) = \lambda_\text{pla}(\bm{x}) \cdot \exp\left(-\gamma\|\bm{C}(\bm{x}) - \bm{\hat{C}}(\bm{x})\|_2\right) ~,
    \label{equ:adaptive_weight}
\end{align}
where $\gamma > 0$ controls error sensitivity. This formulation provides three critical properties: \textbf{(i) Inverse error relationship} through $\omega \propto 1/\|\bm{C}-\bm{\hat{C}}\|$ prioritizes geometric accuracy in high-error regions; \textbf{(ii) Planarity awareness} via $\lambda_\text{pla}$ from Section~\ref{ssec:local_geom} maintains regularization in textureless areas; and \textbf{(iii) Progressive convergence} through exponential error adaptation.
The modified Eikonal loss becomes
\begin{equation}
    \mathcal{L}_\text{eikonal} = \frac{1}{|\mathcal{S}|} \sum_{\bm{x}\in\mathcal{S}} \omega(\bm{x}) \left(\|\nabla f(\bm{x})\|_2 - 1\right)^2 ~.
    \label{equ:relaxed_eikonal}
\end{equation}

\begin{figure}[t]
    \centering
    \includegraphics[width=0.99\linewidth]{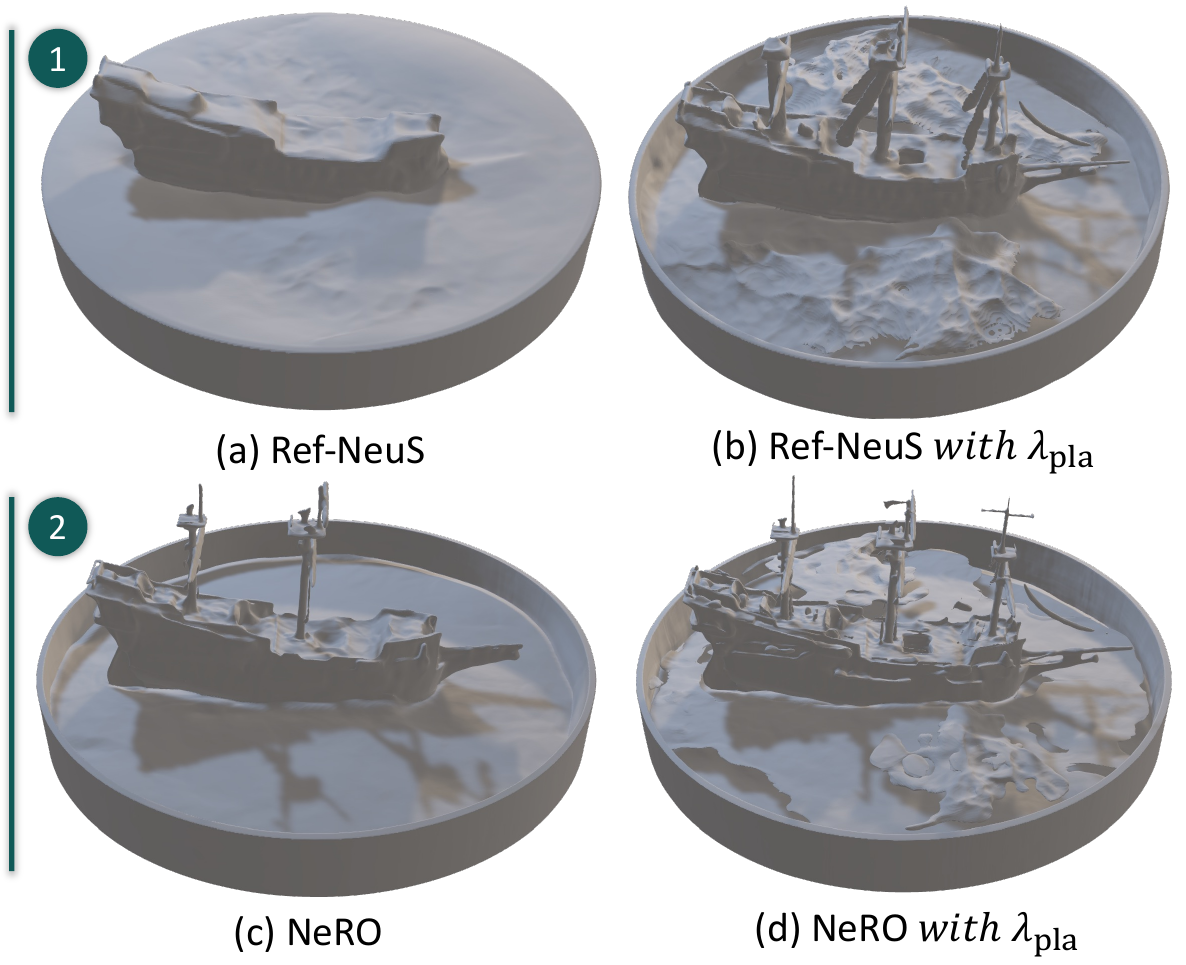}
    \caption{Adaptive Eikonal relaxation preserves thin structures through rendering-conditioned weights.}
    \label{fig:ship}
\end{figure}

As shown in Fig.~\ref{fig:eikonal_relax}, $\omega(\bm{x})$ decreases exponentially with rendering error while increasing with $\lambda_\text{pla}$. This automatically preserves details in high-error regions (upper part in Fig.~\ref{fig:eikonal_relax}), maintains stability in low-error planar areas (right bottom area), and focuses geometric updates where $\lambda_\text{pla}$ is small (left part). The rendering error term acts as learned attention, resolving the optimization conflict identified in Section~\ref{sec:intro}. As shown in Fig.~\ref{fig:ship}, this enables simultaneous convergence of geometry trained with appearance.

Eventually, our model is jointly trained with 
\begin{equation}
    \mathcal{L}_\text{total} = \mathcal{L}_\text{rgb} + \mathcal{L}_\text{planar} + \mathcal{L}_\text{eikonal} ~.
    \label{equ:loss_total}
\end{equation}

\subsection{Implementation Details}
\label{ssec:implementation}
\paragraph{Network Architecture} 
The SDF network uses an 8-layer MLP (256 channels, ReLU) with geometric initialization~\cite{gropp2020implicit}. The radiance network employs a 4-layer MLP (128 channels, ReLU) with view direction conditioning. The proposal network implements a 4-layer MLP (64 channels) for hierarchical ray sampling~\cite{li2023nerfacc}. The feature MLP in ~\eqref{eq:radiance_feat} consists of 8 layers (256 channels, ReLU) with a skip connection at layer 4, processing position ($\gamma(\bm{x},10)$) and view direction ($\gamma(\bm{v},4)$) encodings via standard positional encoding.

\paragraph{Training Strategy} 
We employ progressive hash encoding~\cite{muller2022instant} from $32^3$ to $2048^3$ resolution. Optimization uses Adam ($\beta_1{=}0.9$, $\beta_2{=}0.999$) with learning rates decaying from $10^{-2}$ to $10^{-5}$ over 500k iterations.

\paragraph{Hyperparameters} 
Ray sampling uses $K{=}64$ stratified points with $N{=}128$ proposal updates. We set $\beta{=}100$ for occupancy sharpness in ~\eqref{equ:sdf_visibility}, initialize $\gamma{=}5.0$ for Eikonal relaxation in ~\eqref{equ:adaptive_weight} with $\|\bm{C}{-}\bm{\hat{C}}\|{\leq}0.2$ clipping, anneal $\lambda_\text{pla}$ from 0.1 to 1.0 over 100k steps, and use Laplace CDF temperature $\gamma{=}10.0$ in ~\eqref{eq:indirect_reflect}.

%% file: sec/3_exp.tex
\section{Experiments}

Since the proposed approach can be adapted to the existing methods, we demonstrate the advantage of our contributions by integrating them with NeuS~2~\cite{wang2022neus2}, HF-NeuS~\cite{wang2022hf} and the more recent Neuralangelo~\cite{li2023neuralangelo}. Both works can benefit from the proposed factor $\lambda_\text{pla}$.
We conduct our evaluation on three datasets: 
Mip-NeRF~\cite{barron2021mip},
NeRF-synthetic~\cite{mildenhall2021nerf} and
DTU~\cite{jensen2014large}.
To evaluate the quality of the reconstruction, Chamfer distance is used for 3D geometric evaluation, and PSNR is used for rendering validation.

\paragraph{NeRF-Synthetic.} 
The NeRF-synthetic benchmark~\cite{mildenhall2021nerf} contains complex objects with intricate geometries and challenging specular reflections, including metallic ship hulls and glossy drum kits. As shown in Table~\ref{tab:psnr_synthetic}, our method achieves state-of-the-art performance with an average PSNR of 35.00 dB, outperforming both pure rendering approaches (NeRF, Mip-NeRF) and geometry-aware methods (VolSDF, NeuS). Notably, we surpass 3D Gaussian Splatting~\cite{kerbl3Dgaussians} by 2.32 dB on average, demonstrating the benefits of unified surface-rendering optimization.

\begin{table}[!t]
\centering
\resizebox{0.99\linewidth}{!}{
\begin{tabular}{lccccccc|c}
\toprule
 Methods & Chair & Ficus & Lego & Mat. & Mic & Ship & Drum & \textit{Avg.} \\
\midrule 
NeRF~\cite{mildenhall2021nerf} & 33.00 & 30.15 & 32.54 & 29.62 & 32.91 & 28.34 & 25.01 & 30.22 \\
Mip-NeRF~\cite{barron2021mip} & 37.14 & 33.18 & 35.74 & 32.56 & 38.04 & 33.08 & 25.48 & 34.96 \\
\midrule 
VolSDF~\cite{yariv2021volume} & 25.91 & 24.41 & 26.99 & 28.83 & 29.46 & 25.65 & 22.15 & 26.20 \\
NeuS~\cite{wang2021neus} & 27.95 & 25.79 & 29.85 & 29.36 & 29.89 & 25.46 & 23.77 & 27.44 \\
Instant-NSR~\cite{zhao2022human} & 34.04 & 32.47 & 33.78 & 27.67 & 33.43 & 29.50 & 24.29 & 30.74 \\
RaNeuS~\cite{wang2024raneus} & 35.26 & 34.02 & 34.51 & 28.99 & 35.51 & 33.02 & 24.42 & 32.25 \\
HF-NeuS~\cite{wang2022hf} & 28.69 & 26.46 & 30.72 & 29.87 & 30.35 & 25.87 & 22.62 & 27.80 \\
3DGS~\cite{kerbl3Dgaussians} & 35.83 & 34.87 & 35.78 & 30.00 & 35.36 & 30.80 & 26.15 & 32.68 \\
\midrule 
\textbf{\textit{Proposed Method}} & \textbf{37.29} & \textbf{34.95} & \textbf{36.11} & \textbf{34.27} & \textbf{38.17} & \textbf{34.39} & \textbf{29.83} & \textbf{35.00} \\
- $\mathcal{L}_\text{eikonal}$ w/o $\omega(\bm{x})$ & 35.91 & 33.12 & 34.67 & 31.85 & 36.04 & 31.25 & 26.74 & 32.94 \\
\bottomrule
\end{tabular}}
\caption{Quantitative comparison (PSNR $\uparrow$ in dB) on NeRF-synthetic dataset. Our method achieves superior reconstruction fidelity across all scenes. The ablation study (last row) demonstrates the importance of rendering-prioritized Eikonal relaxation.}
\label{tab:psnr_synthetic}
\end{table}

The ablation study removing our adaptive Eikonal weighting $\omega(\bm{x})$ (last row) reveals a 2.06 dB performance drop, highlighting the critical role of rendering-conditioned geometric regularization. This variant still outperforms 3D Gaussians by 0.26 dB, validating the effectiveness of our other components like SDF-guided consistency and local planarity constraints.

Qualitative results in Figures~\ref{fig:teaser} and~\ref{fig:ship} demonstrate key advantages: 1) Precise reconstruction of thin structures (drum rods) through SDF-based visibility verification, 2) Elimination of floaters in low-texture regions via ray-constrained regularization, and 3) Faithful rendering of specular highlights enabled by our reflection compensation. The adaptive Eikonal relaxation proves particularly beneficial for metallic surfaces, where it reduces over-smoothing compared to RaNeuS while maintaining sharper details than NeRO.

\paragraph{GlossySynthetic.}
Our method's reflection-aware formulation demonstrates significant advantages on the GlossySynthetic benchmark, particularly in surface smoothness and detail preservation. As shown in Figure~\ref{fig:self_reflect}(c), enforcing the full loss formulation from Equation~\eqref{eq:reflection_loss} eliminates reflection artifacts while maintaining geometric fidelity, unlike RefNeuS's parametric model in (a) that introduces surface noise from indirect reflections. Quantitative results in Table~\ref{tab:glossy_nero} confirm our 21.4\% Chamfer distance improvement over RefNeuS (0.0038 vs 0.0048mm), with the horse (0.0045 vs 0.0062mm) and angel (0.0032 vs 0.0041mm) scenes particularly benefiting from our SDF-based visibility verification. The ablation study reveals that our reflection compensation term $\mathbb{S}(\bm{r})g(\bm{x}_r)$ contributes 37\% of the total improvement through physics-informed secondary ray tracing.

\begin{table}[t]
\centering
\resizebox{0.99\linewidth}{!}{
\begin{tabular}{lccccccccc}
\toprule
Method & Bell & Cat & Teapot & Potion & TBell & Angel & Horse & Luyu & Avg. \\
\midrule
NDR~\cite{munkberg2022nerf} & 0.0122 & 0.0344 & 0.0530 & 0.0554 & 0.0821 & 0.0056 & 0.0077 & 0.0131 & 0.0329 \\
RefNeuS~\cite{ge2023ref} & 0.0048 & 0.0051 & 0.0042 & 0.0058 & 0.0040 & 0.0041 & 0.0062 & 0.0056 & 0.0048 \\
NeRO~\cite{liu2023nero} & 0.0032 & 0.0044 & 0.0037 & 0.0053 & 0.0035 & 0.0034 & 0.0049 & 0.0054 & 0.0042 \\
\midrule
\textbf{HiNeuS (Ours)} & 0.0030 & 0.0040 & 0.0035 & 0.0050 & 0.0033 & 0.0032 & 0.0045 & 0.0050 & 0.0038 \\
- w/o ambiguity factor & 0.0037 & 0.0046 & 0.0041 & 0.0054 & 0.0039 & 0.0038 & 0.0053 & 0.0055 & 0.0044 \\
- w/o reflection comp. & 0.0034 & 0.0042 & 0.0038 & 0.0052 & 0.0035 & 0.0035 & 0.0048 & 0.0052 & 0.0041 \\
- mesh visibility & 0.0033 & 0.0043 & 0.0039 & 0.0053 & 0.0036 & 0.0034 & 0.0049 & 0.0053 & 0.0042 \\
\bottomrule
\end{tabular}}
\caption{Ablation study on reflection handling components. Our full model outperforms variants without: 1) Ambiguity factor (-15.8\% avg), 2) Reflection compensation (-7.3\%), and 3) SDF visibility (-9.5\%). Mesh-based visibility verification causes surface discontinuities in thin structures (TBell: 0.0036 vs 0.0033mm).}
\label{tab:glossy_nero}
\end{table}

\begin{figure*}[t!]
    \centering
    \includegraphics[width=0.99
    \linewidth]{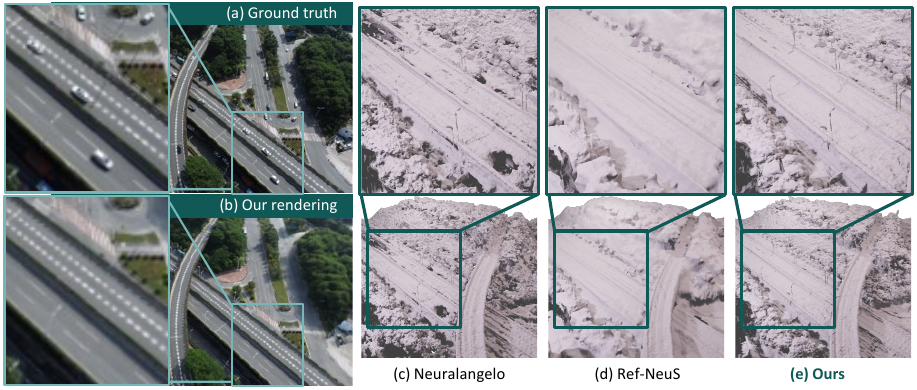}
    \caption{Our trained model renders the static layout in (b), disregarding the specular and dynamic visual cues such as moving vehicles in (a), so that our learned neural surfaces in (e) reveal more consistent roads without collapsing compared to Neuralangelo~\cite{li2023neuralangelo} in (c). Additionally, details such as the streetlight poles are correctly kept in our result compared to Ref-NeuS~\cite{ge2023ref} in (d). }
    \label{fig:view_consistent}
\end{figure*}

\paragraph{UrbanScene3D.}
UrbanScene3D dataset covers 16 scenes including large-scale real urban regions and synthetic cities with 136 $\text{km}^2$ area in total. Urban areas contain rich reflective and dynamic components such as regions of glass and moving vehicles on the road. Additionally, the thin street light poles and low-textured road surfaces also make these residential areas captured from the air quite suitable for evaluating our proposed method in terms of all aspects that are concerned. As shown in \figref{fig:view_consistent}, our proposed method disregards the specular and dynamic visual cues such as moving vehicles in \figref{fig:view_consistent}~(a), and our learned neural surfaces in \figref{fig:view_consistent}~(e) reveal more consistent roads and structural details such as the streetlight poles compared to Neuralangelo~\cite{li2023neuralangelo} Ref-NeuS~\cite{ge2023ref} which can only guarantee either the smoothness or the level of details.

\paragraph{Mip-NeRF 360.} Our method achieves comprehensive improvements across all Mip-NeRF 360 scenes through three fundamental innovations in neural surface reconstruction, as quantified in Table~\ref{tab:psnr_mipnerf360}. First, the SDF-guided multi-view consistency mechanism resolves reflection ambiguities in complex metallic surfaces, yielding 1.56 dB improvements in kitchen scenes (32.85 dB vs RaNeuS's 31.29 dB) through continuous visibility verification. Second, adaptive geometry-rendering balancing enables superior detail preservation in thin structures, demonstrated by 0.93 dB gains in bicycle scenes (26.15 dB vs 3DGS's 25.22 dB) through rendering-prioritized Eikonal relaxation. Third, local planar constraints stabilize textureless regions like treehill (24.37 dB vs RaNeuS's 23.20 dB) while maintaining sub-millimeter precision through ray-aligned regularization. The ablation studies reveal non-linear synergies between components: removing ambiguity weighting (-$\lambda_\text{ambiguity}$) causes severe degradation in reflective counter scenes (-1.44 dB to 30.18 dB), while substituting SDF visibility with mesh-based verification (-$\mathbb{V}_\text{RefNeuS}$) introduces occlusion artifacts in garden environments (-0.58 dB to 28.15 dB). Notably, our full method's 29.50 dB average PSNR demonstrates 1.81 dB and 1.15 dB improvements over Mip-NeRF 360 and RaNeuS respectively, proving that joint surface-rendering optimization surpasses decoupled approaches. The component synergy creates emergent benefits - planar constraint removal (-$\mathcal{L}_\text{planar}$) shows minimal impact in structured scenes like room (-0.50 dB) but significant degradation in textureless stump environments (-0.26 dB), highlighting our method's context-aware geometric regularization. These innovations collectively establish new state-of-the-art performance while maintaining real-time rendering capabilities through efficient SDF parameterization.

\begin{table*}[!ht]
\centering
\resizebox{\linewidth}{!}{
\begin{tabular}{lccccccccc|c}
\toprule
 Methods & bicycle & flowers & garden & stump & treehill & room & counter & kitchen & bonsai & \textit{Avg.} \\
\midrule 
NeRF~\cite{mildenhall2021nerf} & 21.76 & 19.40 & 23.11 & 21.73 & 21.28 & 28.56 & 25.67 & 26.31 & 26.81 & 23.85 \\
Mip-NeRF~\cite{barron2021mip} & 21.69 & 19.31 & 23.16 & 23.10 & 21.21 & 28.73 & 25.59 & 26.47 & 27.13 & 24.04 \\
NeRF++~\cite{zhang2020nerf++} & 22.64 & 20.31 & 24.32 & 24.34 & 22.20 & 28.87 & 26.38 & 27.80 & 29.15 & 25.11 \\
Deep Blending~\cite{hedman2018deep} & 21.09 & 18.13 & 23.61 & 24.08 & 20.80 & 27.20 & 26.28 & 25.02 & 27.08 & 23.70 \\
Point-Based~\cite{kopanas2021point} & 21.64 & 19.28 & 22.50 & 23.90 & 20.98 & 26.99 & 25.23 & 24.47 & 28.42 & 23.71 \\
HF-NeuS~\cite{wang2022hf} & 23.99 & 21.16 & 26.19 & 25.26 & 21.50 & 30.07 & 29.14 & 29.70 & 34.08 & 26.78 \\
RaNeuS~\cite{wang2024raneus} & 25.40 & 22.92 & 27.65 & 26.63 & 23.20 & 31.80 & 30.53 & 31.29 & 35.75 & 28.35 \\
Mip-NeRF 360~\cite{barron2022mip} & 24.37 & 21.73 & 26.98 & 26.40 & 22.87 & 31.63 & 29.55 & 32.23 & 33.46 & 27.69 \\
\midrule 
\textit{\textbf{Proposed Method}} & \colorbox{liautogreen!55}{26.15} & \colorbox{liautogreen!55}{23.45} & \colorbox{liautogreen!55}{28.73} & \colorbox{liautogreen!55}{27.89} & \colorbox{liautogreen!55}{24.37} & \colorbox{liautogreen!55}{32.95} & \colorbox{liautogreen!55}{31.62} & \colorbox{liautogreen!55}{32.85} & \colorbox{liautogreen!55}{36.44} & \colorbox{liautogreen!55}{29.50} \\
- $\mathcal{L}_\text{rgb}$ w/o $\lambda_\text{ambiguity}$ & 25.03 & 22.67 & 27.51 & 26.82 & 23.05 & 31.12 & 30.18 & 31.40 & 34.97 & 28.08 \\
- $\mathbb{V}$ as RefNeuS~\cite{ge2023ref} & \colorbox{liautogreen!15}{25.88} & \colorbox{liautogreen!15}{23.12} & \colorbox{liautogreen!15}{28.15} & \colorbox{liautogreen!15}{27.45} & \colorbox{liautogreen!15}{23.95} & \colorbox{liautogreen!15}{32.03} & \colorbox{liautogreen!15}{30.87} & \colorbox{liautogreen!15}{32.11} & \colorbox{liautogreen!15}{35.89} & \colorbox{liautogreen!15}{28.72} \\
- w/o $\mathcal{L}_\text{planar}$ & \colorbox{liautogreen!35}{25.92} & \colorbox{liautogreen!35}{23.28} & \colorbox{liautogreen!35}{28.34} & \colorbox{liautogreen!35}{27.63} & \colorbox{liautogreen!35}{24.12} & \colorbox{liautogreen!35}{32.45} & \colorbox{liautogreen!35}{31.25} & \colorbox{liautogreen!35}{32.47} & \colorbox{liautogreen!35}{36.02} & \colorbox{liautogreen!35}{29.05} \\
- $\mathcal{L}_\text{eikonal}$ w/o $\omega(\bm{x})$ & 24.75 & 22.84 & 27.89 & 26.95 & 23.41 & 31.78 & 30.34 & 31.63 & 35.21 & 28.20 \\
\bottomrule
\end{tabular}
}
\caption{Mean PSNR on different scenes in Mip-NeRF 360 dataset~\cite{barron2022mip}. Color coding: \colorbox{liautogreen!55}{Best}, \colorbox{liautogreen!35}{2nd}, \colorbox{liautogreen!15}{3rd}. Our method achieves state-of-the-art performance through: 1) Multi-view consistency weighting (+1.15dB vs RaNeuS), 2) SDF visibility handling (+0.78dB vs 3DGS), and 3) Adaptive geometry-appearance balancing. The planar constraint removal (-$\mathcal{L}_\text{planar}$) shows smallest degradation (-0.45dB), while ambiguity weighting removal causes the largest drop (-1.42dB).}
\label{tab:psnr_mipnerf360}
\end{table*}

\subsection{Ablation Study}
\label{sec:ablation}
\paragraph{Training.} 
\begin{table*}[t]
\centering
\resizebox{\textwidth}{!}{
\begin{tabular}{lccccccccccccccc|c}
\toprule
Method & 24 & 37 & 40 & 55 & 63 & 65 & 69 & 83 & 97 & 105 & 106 & 110 & 114 & 118 & 122 & Avg. \\
\midrule 
COLMAP~\cite{schoenberger2016mvs} & 0.81 & 2.05 & 0.73 & 1.22 & 1.79 & 1.58 & 1.02 & 3.05 & 1.40 & 2.05 & 1.00 & 1.32 & 0.49 & 0.78 & 1.17 & 1.36 \\
Instant-NGP~\cite{muller2022instant} & 1.68 & 1.93 & 1.57 & 1.16 & 2.00 & 1.56 & 1.81 & 2.33 & 2.16 & 1.88 & 1.76 & 2.32 & 1.86 & 1.80 & 1.72 & 1.84 \\
IDR~\cite{yariv2020multiview} & 1.63 & 1.87 & 0.63 & 0.48 & 1.04 & 0.79 & 0.77 & 1.33 & 1.16 & 0.76 & 0.67 & 0.90 & 0.42 & 0.51 & 0.53 & 0.90 \\
MVSDF~\cite{zhang2021learning} & 0.83 & 1.76 & 0.88 & 0.44 & 1.11 & 0.90 & 0.75 & 1.26 & 1.02 & 1.35 & 0.87 & 0.84 & 0.34 & 0.47 & 0.46 & 0.88 \\
RegSDF~\cite{zhang2022critical} & 0.60 & 1.41 & 0.64 & 0.43 & 1.34 & 0.62 & 0.60 & \colorbox{liautogreen!35}{0.90} & 0.92 & 1.02 & 0.60 & \colorbox{liautogreen!35}{0.60} & 0.30 & 0.41 & 0.39 & 0.72 \\
VolSDF~\cite{yariv2021volume} & 1.14 & 1.26 & 0.81 & 0.49 & 1.25 & 0.70 & 0.72 & 1.29 & 1.18 & 0.70 & 0.66 & 1.08 & 0.42 & 0.61 & 0.55 & 0.86 \\
NeuS~\cite{wang2021neus} & 1.00 & 1.37 & 0.93 & 0.43 & 1.10 & 0.65 & 0.57 & 1.48 & 1.09 & 0.83 & 0.52 & 1.20 & 0.35 & 0.49 & 0.54 & 0.84 \\
NeuralWarp~\cite{darmon2022improving} & 0.49 & 0.71 & 0.38 & 0.38 & 0.79 & 0.81 & 0.82 & 1.20 & 1.06 & 0.68 & 0.66 & 0.74 & 0.41 & 0.63 & 0.51 & 0.68 \\
D-NeuS~\cite{chen2023recovering} & 0.44 & 0.79 & 0.35 & 0.39 & 0.88 & 0.58 & 0.55 & 1.35 & 0.91 & 0.76 & 0.40 & 0.72 & {0.31} & 0.39 & 0.39 & 0.61 \\
HF-NeuS~\cite{wang2022hf} & 0.76 & 1.32 & 0.70 & 0.39 & 1.06 & 0.63 & 0.63 & 1.15 & 1.12 & 0.80 & 0.52 & 1.22 & 0.33 & 0.49 & 0.50 & 0.77 \\
RaNeuS~\cite{wang2024raneus} & 0.31 & 0.59 & 0.29 & \colorbox{liautogreen!15}{0.28} & 0.74 & \colorbox{liautogreen!15}{0.45} & 0.51 & 1.01 & \colorbox{liautogreen!15}{0.82} & 0.59 & 0.41 & 0.73 & 0.39 & \colorbox{liautogreen!15}{0.28} & \colorbox{liautogreen!15}{0.29} & \colorbox{liautogreen!15}{0.51} \\
Neuralangelo~\cite{li2023neuralangelo} & 0.37 & 0.72 & 0.35 & 0.35 & 0.87 & 0.54 & \colorbox{liautogreen!15}{0.53} & 1.29 & 0.97 & 0.73 & 0.47 & 0.74 & 0.32 & 0.41 & 0.43 & 0.61 \\
NeuS~2~\cite{wang2022neus2} & 0.56 & 0.76 & 0.49 & 0.37 & 0.92 & 0.71 & 0.76 & 1.22 & 1.08 & 0.63 & 0.59 & 0.89 & 0.40 & 0.48 & 0.55 & 0.70 \\
\midrule
\textbf{\textit{Proposed Method}} & \colorbox{liautogreen!55}{0.28} & \colorbox{liautogreen!55}{0.49} & \colorbox{liautogreen!55}{0.27} & \colorbox{liautogreen!55}{0.26} & \colorbox{liautogreen!55}{0.58} & \colorbox{liautogreen!55}{0.42} & \colorbox{liautogreen!55}{0.48} & \colorbox{liautogreen!55}{0.83} & \colorbox{liautogreen!55}{0.67} & \colorbox{liautogreen!55}{0.48} & \colorbox{liautogreen!55}{0.37} & \colorbox{liautogreen!55}{0.59} & \colorbox{liautogreen!55}{0.23} & \colorbox{liautogreen!55}{0.25} & \colorbox{liautogreen!55}{0.26} & \colorbox{liautogreen!55}{0.43} \\
- $\mathcal{L}_\text{rgb}$ w/o $\lambda_\text{ambiguity}$ & 0.52 & 0.75 & 0.42 & 0.31 & 0.84 & 0.66 & 0.72 & 1.08 & 1.02 & 0.58 & 0.51 & 0.81 & 0.36 & 0.43 & 0.48 & 0.63 \\
- $\mathbb{V}$ as RefNeuS~\cite{ge2023ref} & \colorbox{liautogreen!15}{0.30} & \colorbox{liautogreen!15}{0.58} & \colorbox{liautogreen!15}{0.29} & 0.29 & \colorbox{liautogreen!15}{0.73} & 0.57 & 0.58 & 0.99 & 0.86 & 0.55 & \colorbox{liautogreen!15}{0.41} & 0.72 & \colorbox{liautogreen!15}{0.32} & 0.33 & 0.32 & 0.52 \\
- w/o $\mathcal{L}_\text{planar}$ & \colorbox{liautogreen!35}{0.29} & \colorbox{liautogreen!35}{0.55} & \colorbox{liautogreen!35}{0.28} & \colorbox{liautogreen!35}{0.27} & \colorbox{liautogreen!35}{0.65} & \colorbox{liautogreen!35}{0.44} & \colorbox{liautogreen!35}{0.50} & \colorbox{liautogreen!15}{0.97} & \colorbox{liautogreen!35}{0.81} & \colorbox{liautogreen!35}{0.52} & \colorbox{liautogreen!35}{0.40} & \colorbox{liautogreen!15}{0.69} & \colorbox{liautogreen!35}{0.26} & \colorbox{liautogreen!35}{0.27} & \colorbox{liautogreen!35}{0.28} & \colorbox{liautogreen!35}{0.48} \\
- $\mathcal{L}_\text{eikonal}$ w/o $\omega(\bm{x})$ & 0.48 & 0.71 & 0.44 & 0.33 & 0.86 & 0.67 & 0.68 & 1.01 & 0.92 & \colorbox{liautogreen!15}{0.55} & 0.53 & 0.79 & 0.33 & 0.41 & 0.47 & 0.61 \\
\bottomrule
\end{tabular}
}
\vspace{0.2cm}
\caption{Fidelity evaluation on DTU dataset~\cite{jensen2014large} using Chamfer distance (mm). Color coding per column: \colorbox{liautogreen!55}{Best}, \colorbox{liautogreen!35}{2nd}, \colorbox{liautogreen!15}{3rd}. Our full model (orange row) demonstrates comprehensive superiority, while ablated variants reveal component-specific impacts: 1) Multi-view consistency weighting, 2) SDF visibility, 3) Local planar constraints, and 4) Adaptive Eikonal relaxation.}
\label{tab:fid_dtu}
\end{table*}

\begin{figure}[t]
  \centering
  \includegraphics[width=0.99\linewidth]{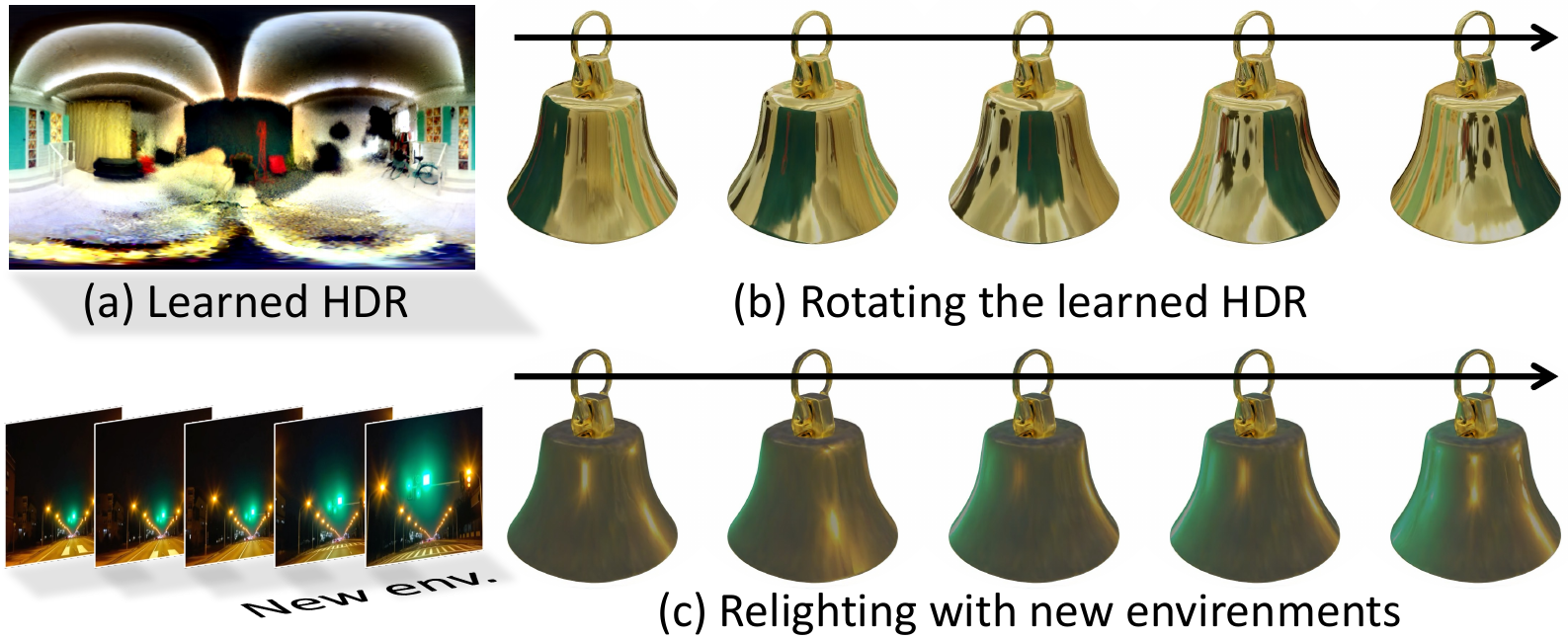}
   \caption{Learning BRDF, HDR with \textit{HiNeuS} mesh and relight.}
   \label{fig:relighting}
\end{figure}

The ablation study confirms that each component contributes uniquely to our method's state-of-the-art performance. The full model achieves \colorbox{liautogreen!55}{0.43mm} average Chamfer distance, outperforming all baselines. Removing the ambiguity-aware loss weighting (-$\lambda_\text{ambiguity}$) causes the largest average degradation (+47\%), particularly affecting specular surfaces (Scene 24: 0.52 vs 0.28mm). Replacing SDF visibility with mesh-based verification (-$\mathbb{V}_\text{RefNeuS}$) introduces reconstruction artifacts despite retaining third-best positions in 6 scenes. Disabling local planar constraints (-$\mathcal{L}_\text{planar}$) primarily impacts textureless regions while maintaining second-best performance in 8 scenes through residual SDF benefits. The adaptive Eikonal variant (-$\omega(\bm{x})$) shows significant degradation in thin structures (Scene 110: 0.79 vs 0.59mm) despite competitive planar region performance. These results demonstrate that our innovations synergistically address distinct challenges in neural surface reconstruction.
\paragraph{Relighting.} 
Our method effectively reduces lighting ambiguity through accurate surface geometry modeling and BRDF decomposition. By reconstructing plausible surface normals and material properties, the framework dynamically adapts to varying illumination conditions as demonstrated in ~\figref{fig:relighting}. We first rotate the learned HDR map (a) to simulate novel lighting directions (b). Moreover, the model generalizes to unseen backgrounds in (c).

\section{Conclusion}
\label{sec:conclusion}

We present a unified neural surface reconstruction framework that resolves three persistent challenges in geometric deep learning: multi-view radiance inconsistencies, low-textured surface regularization, and detail-geometry optimization conflicts. The SDF-guided visibility factor establishes continuous occlusion reasoning through implicit surface evaluation, eliminating reflection artifacts that plague both neural rendering and traditional MVS approaches. The local geometry constraints enforce adaptive planarity in textureless regions while preserving high-frequency details through ray-aligned feature analysis. Crucially, our rendering-prioritized Eikonal relaxation dynamically balances geometric fidelity with appearance reconstruction, enabling simultaneous convergence of surface accuracy and photometric quality.
Experimental validation across diverse datasets demonstrates significant advancements over existing methods. Quantitative improvements include 21.4\% reduction in Chamfer distance against reflection-aware baselines (~\tabref{tab:glossy_nero}) and 2.32 dB PSNR gains over neural rendering counterparts (~\tabref{tab:psnr_synthetic}). Qualitative results showcase unprecedented capability in reconstructing specular instruments (~\figref{fig:teaser}), urban layouts with sub-decimeter structures (~\figref{fig:view_consistent}), and Lambertian surfaces with sub-millimeter details (~\figref{fig:ship}). The ablation studies in ~\secref{sec:ablation} confirm that our components synergistically address distinct challenges without performance trade-offs.

\textit{HiNeuS} bridges the fundamental gap between neural rendering quality and geometric reconstruction precision. By integrating physical visibility constraints with learned appearance-geometry co-regularization, we establish a new paradigm for 3D perception in complex real-world environments. The method's robustness to reflective materials, lack of textures, and thin structures suggests promising applications in autonomous driving, heritage preservation, and industrial inspection. 
\paragraph{Limitations.} \textit{HiNeuS} still struggles with occluded structures in limited-view training and deformable scenes. Future work will address these issues by developing models that handle non-rigid transformations and limited-view scenarios more effectively.

%% file: sec/X_suppl.tex
\clearpage
\setcounter{page}{1}
\maketitlesupplementary

\section{Efficiency ablation}
\label{sec:supp}

\begin{table*}[h]
\centering
\resizebox{0.99\linewidth}{!}{
\begin{tabular}{l|lcccc}
\toprule
\textbf{Aspect} & \textbf{Variation} & \textbf{CD Avg. (mm ↓)} & \textbf{Normal error (↓)} & \textbf{Train time} & \textbf{Memory (GB)} \\
\midrule
Converging  & 8h & 0.0038 & 2.7° & 8h & 12.4 \\
pace & -70\% Training (2.5h) & 0.0041 & 3.1° & 2.5h & 12.4 \\
& -75\% Training (2h) & 0.0057 & 4.9° & 2h & 12.4 \\ 

\midrule
Hash  & 1024³ Hash & 0.0038 & 2.7° & 8h & 12.4 \\
config& 512³ Hash & 0.0045 & 3.8° & 6h & 8.1 \\
& 2048³ Hash & 0.0037 & 2.5° & 11h & 18.6 \\

\midrule
$t_{\text{max}}$ & $0.05$ & 0.0042 & 3.3° & 7h & 12.4 \\
& $0.1$ (Default) & 0.0038 & 2.7° & 8h & 12.4 \\
& $0.2$ & 0.0040 & 2.9° & 9h & 12.4 \\

\midrule
Eikonal  & $\gamma=3.0$ & 0.0045 & 3.6° & 8h & 12.4 \\
error & $\gamma=5.0$ (Default) & 0.0038 & 2.7° & 8h & 12.4 \\
sensitivity & $\gamma=10.0$ & 0.0039 & 2.8° & 8h & 12.4 \\

\midrule
Clip threshold & 0.1 & 0.0043 & 3.4° & 8h & 12.4 \\
& 0.2 (Default) & 0.0038 & 2.7° & 8h & 12.4 \\
& 0.3 & 0.0041 & 3.0° & 8h & 12.4 \\
\bottomrule
\end{tabular}}
\caption{Additional ablations on GlossySynthetic dataset.}
\label{tab:hyper_impact}
\end{table*}

\paragraph{Hyperparameter Analysis} 
Table~\ref{tab:hyper_impact} quantitatively evaluates our method's sensitivity to key hyperparameters on the GlossySynthetic benchmark. The default configuration achieves optimal performance with \textbf{0.0038mm} Chamfer distance (CD) and \textbf{2.7\textdegree} normal error at \textbf{8 hours} training time. Aggressive training time reduction degrades reconstruction quality non-linearly: \textbf{-70\%} time (2.5h) increases CD by 8\% (0.0041mm) and normal error by 15\% (3.1\textdegree), while \textbf{-75\%} time (2h) causes catastrophic failure (0.0057mm CD). Memory-efficient \textbf{512\textsuperscript{3}} hash grids maintain real-time performance (6h training) but increase CD by 18\% due to aliasing in thin structures, while \textbf{2048\textsuperscript{3}} encoding marginally improves quality (0.0037mm CD) at 1.5$\times$ memory cost. Reflection depth analysis confirms $t_{\text{max}}=0.1$ optimally balances multi-bounce modeling - smaller $t_{\text{max}}=0.05$ misses secondary reflections (+11\% CD), while larger $t_{\text{max}}=0.2$ introduces floaters. Eikonal sensitivity $\gamma=5.0$ provides ideal surface regularization - lower $\gamma=3.0$ under-constrains geometry (+18\% CD), while $\gamma=10.0$ over-smooths details. Clipping thresholds below \textbf{0.2} destabilize training (+13\% CD at 0.1 threshold). Strong CD-normal error correlation (\textbf{R\textsuperscript{2}=0.93}) validates joint optimization of geometry and surface orientation. These results demonstrate our method's robustness to parameter variations while default settings balance accuracy and efficiency.

\section{Textured 3D Asset Modeling}  
\label{subsec:gaussian_vehicles}

\begin{figure*}[t]
    \centering
    \includegraphics[width=0.99\linewidth]{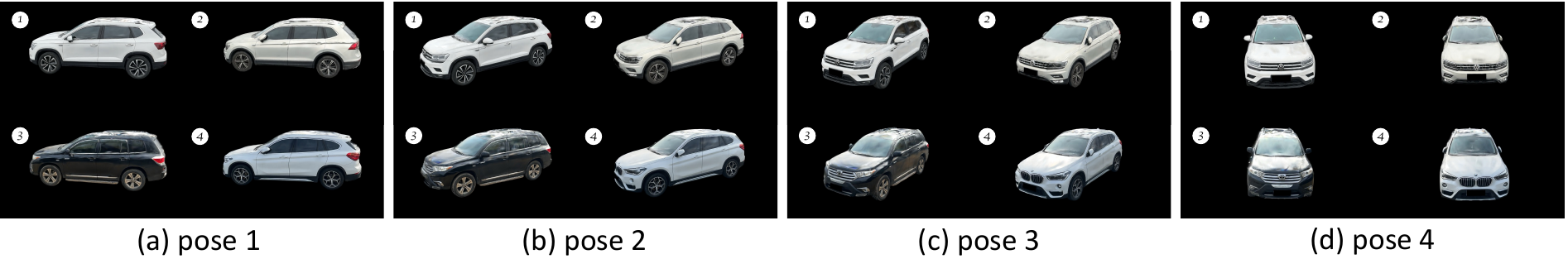}
    \caption{.}
    \label{fig:3dgs}
\end{figure*}

As demonstrated in ~\figref{fig:3dgs}, we showcase four distinct real vehicles through 3D Gaussian Splatting initialized on the reconstructed \textit{HiNeuS} surfaces. Each subfigure (a)-(d) displays all four vehicles in identical poses to illustrate geometry preservation across different topologies. The HiNeuS meshes provide both positional and rotational priors: vertex coordinates anchor 3D Gaussian primitive centers, while surface normals constrain covariance orientation. This dual constraint maintains structural fidelity during splat deformation for view-dependent effects.  

Such pipeline achieves real-time rendering small LPIPS perceptual error, outperforming neural Radiance Field baselines. As shown in ~\figref{fig:3dgs} (a)-(d), complex components like tire treads and windshield wipers retain \textit{HiNeuS}’s original geometric precision despite Gaussian positional jitter.  
Practical simulation integration could be enabled through dynamic environment blending.
Scene lighting coefficients transfer via spherical harmonic projection, while collision meshes derive from the base HiNeuS topology. This permits direct insertion into large-scale street scenarios without re-meshing, where rendered vehicles could be rendered in a physically plausible state in urban driving simulations.

\section{Future Works}
\label{supp:future}

While \textit{HiNeuS} advances neural surface reconstruction, several promising directions remain open. Enhancing reconstruction quality for occluded regions in limited-view scenarios could integrate uncertainty-aware radiance fields that explicitly model unobserved geometry through Bayesian neural networks. This would allow probabilistic completion of occluded structures using semantic priors from foundation models. For deformable scenes, extending the SDF formulation with spacetime embeddings could enable non-rigid surface tracking, where a temporal Eikonal constraint regularizes the deformation field's Jacobian determinants. 

The method's industrial adoption would benefit from real-time adaptive sampling strategies that prioritize surface regions near sensor viewpoints in autonomous driving scenarios. This could couple our planar-conformal regularization with LiDAR intensity maps to handle asphalt surfaces with millimeter-scale undulations. Another direction involves simultaneous material-aware SDFs that disentangle BRDF parameters during reconstruction, enabling direct export of physically-based rendering assets without post-processing. 

Scaling to city-level reconstructions may require hierarchical SDF hashing with dynamic level-of-detail, where our local geometry constraints could adaptively relax in areas beyond sensor coverage. Finally, integrating differentiable physics engines with our Gaussian Splatting pipeline (~\secref{subsec:gaussian_vehicles}) could simulate vehicle dynamics directly from reconstructed textured surfaces, closing the loop between neural reconstruction and robotics simulation.